\def\year{2020}\relax
\documentclass[letterpaper]{article} % DO NOT CHANGE THIS
\usepackage{aaai20}  % DO NOT CHANGE THIS
\usepackage{times}  % DO NOT CHANGE THIS
\usepackage{helvet} % DO NOT CHANGE THIS
\usepackage{courier}  % DO NOT CHANGE THIS
\usepackage[hyphens]{url}  % DO NOT CHANGE THIS
\usepackage{graphicx} % DO NOT CHANGE THIS
\usepackage{graphicx}  % DO NOT CHANGE THIS
\usepackage{amsmath}
\usepackage{amssymb}
\usepackage{makecell}
\title{Story Realization: Expanding Plot Events into Sentences}
\author{Prithviraj Ammanabrolu, Ethan Tien, Wesley Cheung, \\ \bf \Large Zhaochen Luo, William Ma, Lara J. Martin, and Mark O. Riedl \\
School of Interactive Computing\\
Georgia Institute of Technology\\
  \texttt{\{raj.ammanabrolu, etien, wcheung8,}
  \texttt{zluo, wma61, ljmartin, riedl\}@gatech.edu}}
\begin{document}
\maketitle
%\tracingall
\begin{abstract}
%We explore the problem of guided neural language generation in the context of automated story plot generation.
%Automated story generation systems that use neural language models tend to suffer from issues related to plots coherence and a lack of a sense of progression toward a specific goal or ending.
%Automated story plot generation is the problem of generating the important plot points of a story. 
Neural network based approaches to automated story plot generation attempt to learn how to generate novel plots from a corpus of natural language plot summaries.
Prior work has shown that a semantic abstraction of sentences called {\em events} improves neural plot generation and and allows one to decompose the problem into: (1)~the generation of a sequence of events (event-to-event) and (2)~the transformation of these events into natural language sentences (event-to-sentence).
%The translation from events to sentences must restore details essential to human readability but are glossed over during event-to-event generation. 
However, typical neural language generation approaches to event-to-sentence can ignore the event details and produce grammatically-correct but semantically-unrelated sentences.
We present an ensemble-based model that generates natural language guided by events.
%Our method outperforms the baseline sequence-to-sequence model. % on the event-to-sentence problem.
We provide results---including a human subjects study---for a full end-to-end automated story generation system showing that our method generates more coherent and plausible stories than baseline approaches~\footnote{Code to reproduce our experiments is available at~\url{https://github.com/rajammanabrolu/StoryRealization}}.

%The automated storytelling systems that we focus on are capable of learning everything they need from a given textual story corpora, usually using sequence-to-sequence models.
%Prior work has shown that a mid-level semantic abstraction between words and sentences in the form of events enables us to decompose the problem into: the generation of a sequence of events (event-to-event) and the transformation of these events into natural language sentences (event-to-sentence).
%Sequence-to-sequence models, when applied to the event-to-sentence problem, however, work as language models that ignore the given event due to the sparsity of the dataset.
%We present an ensemble based model that provides a method for a more controlled system, that generates natural language guided by the given event.
%Our method outperforms the baseline sequence-to-sequence model on the event-to-sentence problem.
%Additionally, we provide results for a full end-to-end automated story generation system, demonstrating how our model works with existing systems designed for the event-to-event problem.
\end{abstract}

\section{Introduction}

Automated story plot generation is the problem of creating a sequence of main plot points for a story in a given domain.
Generated plots must remain consistent across the entire story, preserve long-term dependencies, and make use of commonsense and schematic knowledge~\cite{Wiseman2017}.
Early work focused on symbolic planning and case-based reasoning ~\cite{Meehan1977,turner1986thematic,Lebowitz1987,PerezyPerezMikeSharples2001,Gervas2005,Porteous2009,Riedl2010a,ware11,Farrell2019} at the expense of manually-engineering world domain knowledge.

In contrast, neural-based approaches to story and plot generation train a neural language model on a corpus of stories to predict the next character, word, or sentence in a sequence based on a history of tokens~\cite{jain2017story,Clark2018a,Fan2018,Martin2018,Peng2018,Roemmele2018}.
The advantage of neural-based approaches is that there is no need for explicit domain modeling beyond providing a corpus of example stories.
The primary pitfall of neural language model approaches for story generation is that the space of stories that can be generated is huge, which in turn, implies that, in a textual story corpora, any given sentence will likely only be seen once.

\citeauthor{Martin2018} (\citeyear{Martin2018}) propose the use of a semantic abstraction called an {\em event}, reducing the sparsity in a dataset that comes from an abundance of unique sentences.
They define an event to be a unit of a story that creates a change in the story world's state.
Technically, an event is a tuple containing a subject, verb, direct object, and some additional disambiguation token(s).

The event representation enables the decomposition of the plot generation task into two sub-problems: {\em event-to-event} and {\em event-to-sentence}.
Event-to-event is broadly the problem of generating the sequence of events that together comprise a plot.
Models used to address this problem are also responsible for maintaining plot coherence and consistency.
Once new events are generated, however, they are still not human-readable.
%These events are abstractions and aren't human-readable.
Thus the second sub-problem, event-to-sentence, focuses on transforming these events into natural language sentences.

\citeauthor{Martin2017a}~\shortcite{Martin2017a,Martin2018} further propose that this latter, event-to-sentence problem can be thought of as a translation task---translating from the language of events into natural language.
We find, however, that the sequence-to-sequence LSTM networks~\cite{Sutskever2014} that they used frequently ignore the input event and only generate text based on the original corpus, overwriting the plot-based decisions made during event-to-event.
There are two contributing factors.
Firstly, event-to-event models tend to produce previously-unseen events, which, when fed into the event-to-sentence model result in unpredictable behavior.
A basic sequence-to-sequence model is unable to learn how to map these unseen events to sentences.
%The mapping from an unseen event to a sentence is unknown to a basic sequence-to-sequence model.
Secondly, sentences are often only seen once in the entire corpus.
Despite the conversion into events, the sparsity of the data means that each event is still likely seen a limited number of times.
For these reasons, we  framed the event-to-sentence task as {\em guided language generation}, using a generated event as a guide.
%\pagebreak

The contributions of this paper are twofold.
We present an ensemble-based system for the event-to-sentence problem that balances between retaining the event's original semantic meaning, while being an interesting continuation of the story.
We demonstrate that our system for guided language generation outperforms a baseline sequence-to-sequence approach.
Additionally, we present the results of a full end-to-end story generation pipeline (Figure~\ref{fig:pipeline}), showing how all of the sub-systems can be integrated.

%The rest of this paper is organized as follows. 
%We first discuss related work on automated story generation. 
%We give details on our version of the event representation and the event-to-event model.
%Next, background knowledge on event generation, our dataset, and event-to-event are described.
%Event-to-sentence is then introduced along with the components of the ensemble as well as the structure of the ensemble itself, followed by results and discussion for both event-to-sentence and the full end-to-end story generation pipeline.

%%%%%%%%%%%%%%%%%%%%%%%%%%%%%%%%%%%

\section{Related Work and Background}

% Early storytelling systems were based on symbolic planning~\cite{PerezyPerezMikeSharples2001,Riedl2010a, Meehan1977,Lebowitz1987,ware11} and case-based reasoning~\cite{turner1986thematic,PerezyPerezMikeSharples2001,Gervas2005}.
% These systems required a high knowledge-engineering overhead in terms of operators or stories transcribed into symbolic form.
% Consequently, these systems were only capable of generating stories in relatively limited domains.
%Although these systems were capable of maintaining long term coherence, they are not robust and incapable of generating stories outside their engineered domains.
%

\subsection{Story Generation via Machine Learning}

Machine learning approaches to story and plot generation attempt to learn domain information from a corpus of story examples~\cite{swanson12,Li}.
Recent work has looked at using recurrent neural networks (RNNs) for story and plot generation.
\citeauthor{Roemmele2018b}~\shortcite{Roemmele2018} use LSTMs  %~\cite{Hochreiter1997}
with skip-though vector embeddings~\cite{kiros2015skip} to generate stories.
\citeauthor{Khalifa2017}~\shortcite{Khalifa2017} train an RNN on a highly-specialized corpus, such as work from a single author.
\citeauthor{Fan2018}~\shortcite{Fan2018} introduce a form of hierarchical story generation in which a premise is first generated by the model and then transformed into a passage.
This last example is a form of guided generation wherein a single sentence provides guidance.
Similarly, \citeauthor{Yao}~\shortcite{Yao} decompose story generation into planning out a storyline and then generating a story from it.
Our work differs in that we use the event-to-event process to provide guidance to event-to-sentence.
\citeauthor{ammanabrolupcg}~\shortcite{ammanabrolupcg} look at narrative generation as a form of quest generation in interactive fiction and use a knowledge graph to ground their generative models.

\subsection{Event Representation and Generation}
\label{sec:e2e}

\begin{figure}
  \centering
  %\fbox{\rule[-.5cm]{0cm}{4cm} \rule[-.5cm]{4cm}{0cm}}
  \includegraphics[width=0.7\linewidth]{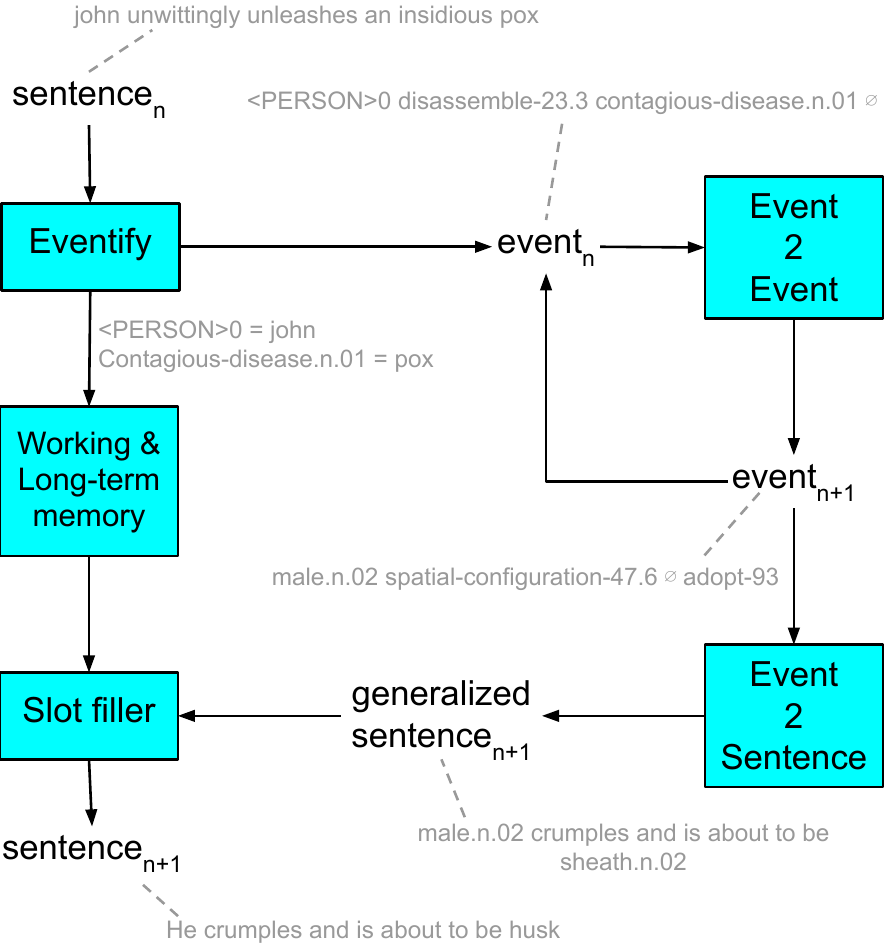}
  \caption{The full automated story generation pipeline,
  %, based off of the proposed pipeline of~\citeauthor{Martin2018}~\shortcite{Martin2018},
  illustrating an example where the event-to-event module generates only a single following event.}
  \label{fig:pipeline}
\end{figure}

% \cite{Martin2017a} propose a system in which the potential methods of solving the event-to-event and event-to-sentence problems are detailed in the context of improvisational storytelling. %, in which a user takes turns with a storytelling agent to tell a story.
% In order to create a full improvisational storytelling pipeline, we first needed to implement an event-to-event model such that generated events can be inputted into our event-to-sentence system. 

\citeauthor{Martin2018}~\shortcite{Martin2018} showed that the performance on both event-to-event and event-to-sentence problems improve when using an abstraction---known as an {\em event}---instead of natural language sentences.
We use a variation of this event structure.
In our work, events are defined as a 5-tuple of $\langle s,v,o,p,m\rangle$ as opposed to the 4-tuples used in \citeauthor{Martin2018}~\shortcite{Martin2018}.
Here $v$ is a verb, $s$ is the subject of the verb, $o$ is the object, $p$ is the corresponding preposition, and $m$ can be a modifier, prepositional object, or indirect object.
Any of these elements can be $\varnothing$, denoting the absence of the element.
All elements are stemmed and generalized with the exception of the preposition.

The generalization process involves finding the VerbNet~\cite{Schuler2005} v3.3 class of the verb and finding the WordNet~\cite{Miller1995} v3.1 Synset that is two levels higher in the hypernym tree for all of the nouns in the event.
This process also includes the identification of named entities in the event tuple, extracting people, organizations, locations, etc. through named entity recognition (NER) and numbering them as the story goes on.
For example ``PERSON'' names are replaced by the tag $<$PERSON$>$$n$ where $n$ indicates the $n$-th ``PERSON'' in the story.
Similarly, the other NER categories are replaced with tags that indicate their category and their number within the story.
This maintains consistency in named entities for a given story in the corpora.

We further process the corpus by ``splitting'' sentences akin to the ``split-and-prune'' methodology of ~\citeauthor{Martin2018}~\shortcite{Martin2018}.
This is done to decrease the number of events generated from a single sentence---reducing the number of mappings of a single sentence to multiple events.
The splitting process starts with extracting the parse trees of each sentence using the Stanford Parser.
Sentences are then split on S's (SBARs) and conjunctions before nested sentences.
This process can result in incomplete sentences where the S-bar phrase is nested inside of a sentence, acting as the direct object. 
For example, when it sees a sentence like ``She says that he is upset.'' it becomes ``She says. He is upset.''
Then the split sentences are sorted to reflect the original ordering of subjects or phrases as closely as possible.

For this paper, the event-to-event system is the policy gradient deep reinforcement learner from \citeauthor{Tambwekar2019}~\shortcite{Tambwekar2019}.
This system has been tested to ensure that the resulting events are of high quality to minimize error in that portion of the pipeline.
Our event-to-sentence system is agnostic to the choice of the event-to-event system, all it requires is a sequence of events to turn into sentences.
The event-to-event network is placed into the pipeline as the ``Event2Event'' module, seen in Figure~\ref{fig:pipeline}, and its output is fed into the event-to-sentence models during testing.

\section{Event-to-Sentence}

We define event-to-sentence to be the problem of selecting a sequence of words $s_t={s_{t_0},s_{t_1},...,s_{t_k}}$---that form a sentence---given the current input event $e_t$, i.e. the current sentence is generated based on maximizing $Pr(s_t|e_t;\theta)$ where $\theta$ refers to the parameters of the generative system.
The eventification in Section~\ref{sec:e2e} is a lossy process in which some of the information from the original sentence is dropped.
Thus, the task of event-to-sentence involves filling in this missing information.
There is also no guarantee that the event-to-event process will produce an event that is part of the event-to-sentence training corpus, simply due to the fact that the space of potentially-generated events is very large; the correct mapping from the generated event to a natural language sentence would be unknown.

In prior work, \citeauthor{Martin2018}~\shortcite{Martin2018} use a sequence-to-sequence LSTM neural network to translate events into sentences.
We observe that ``vanilla'' sequence-to-sequence networks end up operating as simple language models, often ignoring the input event when generating a sentence.
The generated sentence is usually grammatically correct but retains little of the semantic meaning given by the event.

We thus look for other forms of guided neural language generation, with the goals of preserving the semantic meaning from the event in addition to keeping the generated sentences interesting.
We propose four different models---optimized towards a different point in the spectrum between the two objectives, and a baseline fifth model that is used as a fallthrough.
The task of each model is to translate events into ``generalized'' sentences, wherein nouns are replaced by WordNet Synsets.
If a model does not pass a specific threshold (determined individually for each model), the system continues onto the next model in the ensemble.
In order, the models are:
(1)~a retrieve-and-edit model based on~\citeauthor{Hashimoto2018}~\shortcite{Hashimoto2018};
(2)~template filling; 
(3)~sequence-to-sequence with Monte Carlo beam decoding;
(4)~sequence-to-sequence with a finite state machine decoder; and
(5)~vanilla (beam-decoding) sequence-to-sequence.
We find that none of these models by themselves can successfully find a balance between the goals of {\em retaining all of the event tokens} and {\em generating interesting output}.
However, each of the models possess their own strengths and weaknesses---each model is essentially optimized towards a different point on the spectrum between the two goals.
We combine these models into an ensemble in an attempt to minimize the weaknesses of each individual model and to achieve a balance. % between retaining semantic meaning from the event and generating interesting sentences.

\subsection{Retrieve-and-Edit}

The first model is based on the retrieve-and-edit {\em RetEdit} framework for predicting structured outputs~\cite{Hashimoto2018}.
We first learn a task-specific similarity between event tuples by training an encoder-decoder to map each event onto an embedding that can reconstruct the output sentence; this is our retriever model.
Next, we train an editor model which maximizes the likelihood of generating the target sentence given both the input event and a retrieved event-sentence example pair.
We used a standard sequence-to-sequence model with attention and copying \cite{Gu2016} to stand in as our editor architecture.
Although this framework was initially applied to the generation of GitHub Python code and Hearthstone cards, we extend this technique to generate sentences from our event tuples.
Specifically, we first initialize a new set of GLoVe word embeddings~\cite{pennington2014glove}, using random initialization for out-of-vocabulary words.
We use our training set to learn weights for the retriever and editor models, set confidence thresholds for the model with the validation set, and evaluate performance using the test set.

In order to generate a sentence from a given input event, there are two key phases:  ``retrieve'' phase and  ``edit'' phase. 
With respect to the input event, we first retrieve the nearest-neighbor event and its corresponding sentence in the training set using the retriever model. % under the encoder.
%Then,
Passing both the retrieved event-sentence pair and the input event as inputs, we use the editor model to generate a sentence using beam search. 

Many of the successes produced by the model stem from its ability to retain the complex sentence structures that appear in our training corpus and thus attempts to balance between maintaining coherence and being interesting.
However, this interaction with the training data can also prove to be a major drawback of the method; target events that are distant in the embedding space from training examples typically result in poor sentence quality.
Since RetEdit relies heavily on having good examples, we set the confidence of the retrieve-and-edit model to be proportional to $1~$--$~retrieval$ $distance$ when generating sentences, as a lower retrieval distance implies greater confidence.
However, the mapping from event to sentence is not a one-to-one function. There are occasionally multiple sentences that map to a single event, resulting in retrieval distance of 0, in which case the example sentence is returned without  modifications.

\subsection{Sentence Templating}
\label{sec:template}
As mentioned earlier, the baseline sequence-to-sequence network operates as a simple language model and can often ignore the input event when generating a sentence.
However, we know that our inputs, an event tuple will have known parts of speech.%$\langle s,v,p,o,m\rangle$ will have known parts of speech.
%This approach takes advantage of the inputs having the core elements of a basic sentence, $<$$s,v,p,o,m$$>$.
We created a simplified grammar for the syntax of sentences generated from events:
\setlength{\abovedisplayskip}{0pt}
\setlength{\belowdisplayskip}{0pt}
\setlength{\abovedisplayshortskip}{0pt}
\setlength{\belowdisplayshortskip}{0pt}
\begin{align}
%\centering
S &\rightarrow  NP~~~v~~~(NP)~~~(PP)\nonumber\\
NP &\rightarrow  d~~~n \nonumber\\
PP &\rightarrow  p~~~NP \nonumber
\end{align}
\setlength{\abovedisplayskip}{7pt plus2pt minus5pt}
\setlength{\belowdisplayskip}{\abovedisplayskip} 
\setlength{\abovedisplayshortskip}{0pt plus3pt}
\setlength{\belowdisplayshortskip}{4pt plus3pt minus3pt}%
where $d$ is a determiner that will be added and the rest of the terminal symbols correspond to an argument in the event, with $n$ being $s$, $o$, or $m$, depending on its position in the sentence.
The resulting sentence would be
$[\underline{\hspace{3mm}}s] \{v\ [\underline{\hspace{3mm}}o]\ [p\underline{\hspace{3mm}}m]\}$
%$$[\_\_\ s\ \_\_] \{v\ [\_\_\ o\ \_\_]\ [p\ \_\_\ m\ \_\_]\}$$
%It would just be a matter of putting words in places where they belong to fill out the Noun Phrases (NP) and Verb Phrases (VP) to fill out the rest of the sentence.
where blanks indicate where words should be added to make a complete sentence.

First, our algorithm predicts the most likely VerbNet frame based on the contents of the input event (how many and which arguments are filled).
VerbNet provides a number of syntactic structures for different verb classes based on how the verb is being used.
For example, if the input event contains 2 nouns and a verb without a preposition, we assume that the output sentence takes the form of [NP V NP], but if it has 2 nouns, a verb, {\em and} a proposition, then it should be [NP V PP].

Second, we apply a Bidirectional LSTM language model trained on the generalized sentences in our training corpus. 
Given a word, we can generate words before and after it, within a particular phrase as given by some of the rules above, and concatenate the generated sentence fragments together.
%Given a word, we can attempt to generate words in both the backwards and forwards direction, and then concatenate the parts together.
Specifically, we use the AWD-LSTM~\cite{merityRegOpt} architecture as our language model since it is currently state-of-the-art. %has been shown to provide state-of-the-art results for language modeling. 
%We train the language model off of a generalized version of our dataset, so we can attempt to generate words based on the generalized tokens in the input events.

At decode time, we continue to generate words in each phrase until we reach a stopping condition: (1) reaching a maximum length (to prevent run-on sentences); or (2) generating a token that is indicative of an element in the next phrase, for example seeing a verb being generated in a noun phrase. %(seeing a verb generated in a NP).
When picking words from the language model, we noticed that the words ``the'' and ``and'' were extremely common. 
To increase the variety of the sentences, we sample from the top $k$ most-likely next words and enforce a number of grammar-related rules in order to keep the coherence of the sentence.
For example, we do not allow two determiners nor two nouns to be generated next to each other.

One can expect that many of the results will look structurally similar. 
However, we can guarantee that the provided tokens in the event will appear in the generated sentence---this model is optimized towards maintaining coherence. 
To determine the confidence of the model for each sentence, we sum the loss after each generated token, normalize to sentence length, and subtract from 1 as higher loss translates to lower confidence. 

\subsection{Monte-Carlo Beam Search}

Our third method is an adaptation of {\em Monte Carlo Beam Search}~\cite{Cazenave2012} for event-to-sentence. 
We train a sequence-to-sequence model on pairs of events \& generalized sentences and run Monte Carlo beam search at decode time.
This method differs from traditional beam search in that it introduces another scoring term that is used to re-weight all the beams at each timestep. 

After top-scoring words are outputted by the model at each timestep, playouts are done from each word, or {\em node}.
%A node here refers the word from which we begin to perform the playouts.
A node is the final token of the partially-generated sequences on the beam currently and the start of a new playout.
During each playout, one word is sampled from %using the weights given by 
the current step's softmax over all words in the vocabulary.
The decoder network is unrolled until it reaches the ``end-of-story'' tag.
Then, the previously-generated sequence and the sequence generated from the current playout are concatenated together and passed into a scoring function that computes the current playout's score. 

The scoring function is a combination of (1)~BLEU scores up to 4-grams between the input event and generated sentence, as well as (2)~a weighted 1-gram BLEU score between each item in the input event and generated sentence.
The weights combining the 1-gram BLEU scores are learned during validation time where the weight for each word in the event that {\em does not} appear in the final generated sequence gets bumped up. 
Multiple playouts are done from each word and the score $s$ for the current word is computed as:
\begin{equation}
    s_{t} = \alpha * s_{t-1} + \\
     (1 - \alpha) * AVG(playout_{t})
\end{equation}
where $\alpha$ is a constant. 

In the end, the $k$ partial sequences with the highest playout scores are kept as the current beam.
For the ensemble, this model's confidence score is the final score of the highest-scoring end node. 
Monte Carlo beam search excels at creating diverse output---i.e. it skews towards generating interesting sentences. 
Since the score for each word is based on playouts that sample based on weights at each timestep, it is possible for the output to be different across runs. 
The Monte Carlo beam decoder has been shown to generate better sentences that are more grammatically-correct than the other techniques in our ensemble, while sticking more to the input than a traditional beam decoder. 
However, there is no guarantee that all input event tokens will be included in the final output sentence.

\subsection{Finite State Machine Constrained Beams}

Various forms of beam search, including Monte Carlo playouts, cannot ensure that the tokens from an input event appear in the outputted sentence. 
As such, we adapted the algorithm to fit such lexical constraints, similar to \citeauthor{anderson2016guided}~\shortcite{anderson2016guided} who adapted beam search to fit captions for images, with the lexical constraints coming from sets of image tags.
The {\em Constrained Beam Search} used finite state machines to guide the beam search toward generating the desired tokens. 
Their approach, which we have co-opted for event-to-sentence, attempts to achieve a balance between the flexibility and sentence quality typical of a beam search approach, while also adhering to the context and story encoded in the input events that more direct approaches (e.g. Section ~\ref{sec:template}) would achieve.

The algorithm works on a per-event basis, beginning by generating a finite state machine. 
This finite state machine consists of states that % correspond to 
enforce the presence of input tokens in the generated sentence.
As an example, assume we have an $n$-token input event, $\{t_1,t_2,t_3,...,t_n\}$. 
The corresponding machine consists of $2^n$ states. 
Each state maintains a search beam of size $B^s$ with at most $b$ output sequences, corresponding to the configured beam size $s$.
At each time step, every state (barring the initial state) receives from predecessor states those output sequences whose last generated token matches an input event token.
The state then adds to its beam the $b$ most likely output sequences from those received. 
%
%In the example, 
Generating token $t_1$ moves the current state from the initial state to the state corresponding to $t_1$, $t_3$ to a state for $t_3$, and so on. 
The states $t_1$ and $t_3$ then, after generating tokens $t_1$ and $t_3$ respectively, transmit said sequences to the state $t_{1,3}$. 
The states and transitions proceed as such until reaching the final state, wherein they have matched every token in the input event. 
Completed sequences in the final state contain all input event tokens, thus providing us with the ability to retain the semantic meaning of the event.

As much as the algorithm is based around balancing generating good sentences with satisfying lexical constraints, it does not perform particularly well at either. 
It is entirely possible, if not at all frequent, for generated sentences to contain all input tokens but lose proper grammar and syntax, or even fail to %
%generate them all 
reach the final state 
within a fixed time horizon.
This is exacerbated by larger tuples of tokens, seen even at just five tokens per tuple. 
To compensate, we relax our constraint to permit output sequences that have matched at least three out of five tokens from the input event. 

\subsection{Ensemble}
The entire event-to-sentence ensemble is designed as a cascading sequence of models: 
(1)~retrieve-and-edit, (2)~sentence templating, (3)~Monte Carlo beam search, (4)~finite state constrained beam search, and (5)~standard beam search.
We use the confidence scores generated by each of the models in order to re-rank the outputs of the individual models.
This is done by setting a confidence threshold for each of the models such that if a confidence threshold fails, the next model in the ensemble is tried.
The thresholds are tuned on the confidence scores generated from the individual models on the validation set of the corpus.
This ensemble saves on computation as it sequentially queries each model, terminating early and returning an output sentence if the confidence threshold for any of the individual models are met.

An event first goes through the retrieve-and-edit framework, which generates a sentence and corresponding confidence score.
This framework performs well when it is able to retrieve a sample from the training set that is relatively close in terms of retrieval distance to the input.
Given the sparsity of the dataset, this happens with a relatively low probability, and so we place this model first in the sequence.
%The sentence templating approach is the next model that is activated.
%This model is relatively 

The next two models are each optimized towards one of our two main goals.
The sentence templating approach retains all of the tokens within the event and so loses none of its semantic meaning, at the expense of generating a more interesting sentence.
The Monte-Carlo approach, on the other hand, makes no guarantees regarding retaining the original tokens within the event but is capable of generating a diverse set of sentences.
We thus cascade first to the sentence templating model and then the Monte-Carlo approach, implicitly placing greater importance on the goal of retaining the semantic meaning of the event.

The final model queried is the finite-state-machine--constrained beam search.
This model has no confidence score; either the model is successful in producing a sentence within the given length with the event tokens or not.
% and is then taken to the final fall through model.
In the case that the finite state machine based model is unsuccessful in producing a sentence, the final fallthrough model---the baseline sequence-to-sequence model with standard beam search decoding---is used.

\section{Dataset}
\label{sec:scifi}
%\subsection{Sci-fi: A New Dataset}

To aid in the performance of our story generation, we select a single genre: science fiction.
We scraped long-running science fiction TV show plot summaries from the fandom wiki service {\em wikia.com}.
This dataset contains longer and more detailed plot summaries than the dataset used in \citeauthor{Martin2018}~\shortcite{Martin2018} and \citeauthor{Tambwekar2019}~\shortcite{Tambwekar2019}, which we believe to be important for the overall story generation process.
The corpus contains 2,276 stories in total, each story an episode of a TV show.
The average story length is 89.23 sentences.
There are stories from 11 shows, with an average of 207 stories per show, from shows like {\em Doctor Who}, {\em Futurama}, and {\em The X-Files}.
The data was pre-processed to simplify alien names in order to aid the parser.
Then the sentences were split, partially following the ``split-and-pruned'' methodology of \citeauthor{Martin2018}~\shortcite{Martin2018} as described in~\ref{sec:e2e}. 

Once the sentences were split, they were ``eventified'' as described in Section~\ref{sec:e2e}.
One benefit of having split sentences is that there is a higher chance of having a 1:1 correspondence between a sentence and an event, instead of a single sentence becoming multiple events.
After the data is fully prepared, it is split in a 8:1:1 ratio to create the training, validation, and testing sets, respectively.

\section{Experiments}
\begin{table*}[t]
\caption{Event-to-sentence examples for each model. $\varnothing$ represents an empty parameter; $<$PRP$>$ is a pronoun.}
\label{table:examples-e2s}
\scriptsize
\centering
\begin{tabular}{p{1.8cm}|p{2.2cm}|p{2.2cm}|p{1.8cm}|p{2cm}|p{2.8cm}}
{\bf Input Event} & {\bf RetEdit} & {\bf Templates} & {\bf Monte Carlo} & {\bf FSM} & {\bf Gold Standard} \\
\hline
$\langle$$<$PRP$>$, act-114-1-1, to, $\varnothing$, event.n.01$\rangle$& 
$<$PRP$>$ and $<$PERSON$>$0 move to the event$.n.01$ of the natural\_object$.n.01$.&
$<$PRP$>$ act-114-1-1 to event.$n.01$.&
$<$PRP$>$ moves to the nearest natural\_object.$n.01$.&
physical\_entity.$n.01$ move back to the phenomenon.$n.01$ of the craft.$n.02$...&
$<$PRP$>$ move to the event.$n.01$.\\ 
\hline
$\langle$$<$PERSON$>$2, send-11.1, through, $<$PERSON$>$6, $<$LOCATION$>$1$\rangle$& 
$<$PERSON$>$2 sends $<$PERSON$>$6 through the $<$LOCATION$>$1.&
The $<$PERSON$>$2 send-11.1 the $<$PERSON$>$6 through $<$LOCATION$>$1.&
$<$PERSON$>$2 passes this undercover in the body\_part.$n.01$ and collapses.&
$\varnothing$&
In activity.n.01 to avoid $<$PRP$>$ out.n.01 $<$PERSON$>$2 would transport $<$PERSON$>$6 through the $<$LOCATION$>$1.\\
%\hline
%$\langle$$<$PERSON$>$0, admire-31.2, $\varnothing$, $<$PERSON$>$3, $\varnothing$$\rangle$&
%$<$PERSON$>$0 believes $<$PERSON$>$3.&
%$<$PERSON$>$0 admire-31.2 and $<$PERSON$>$3&
%$<$PERSON$>$0 hates $<$PERSON$>$3 saying $<$PRP$>$ s not ready for duration.$n.03$ .& 
%$<$PERSON$>$0 and $<$PERSON$>$0 comes in $<$PRP$>$ content.$n.05$ for wrongdoing.$n.02$ and says %$<$PERSON$>$0 has made on line.$n.23$ have made trait.$n.01$.&
%A pivotal artifact.$n.01$ in $<$PRP$>$ act.$n.02$ is a examination.$n.01$ divised by $<$LOCATION$>$0 to make $<$PERSON$>$0 hate $<$PRP$>$ $<$PERSON$>$3.
\end{tabular}
\end{table*}

\begin{table*}[t]
\caption{End-to-end pipeline examples on previously-unseen input data. The Event-to-Sentence model used is the full ensemble. Sentences are generated using both the extracted and generated events.}% $\varnothing$ represents an empty parameter; $<$PRP$>$ is a pronoun.}
  \label{table:examples-pipe}
\centering
\scriptsize

\begin{tabular}{p{1.8cm}|p{1.4cm}|p{3.4cm}|p{4.6cm}|p{3.6cm}}

{\bf Input Sent.} &{\bf Extracted event} & {\bf Generated Events (Event-to-Event)} & {\bf Generated Sentences (Event-to-Sentence)}&{\bf Slot-filled Sentences} \\
\hline
On Tatooine, Jabba the Hutt inspects the drone barge recently delivered to him.&
$\langle$$<$ORG$>$0, assessment-34.1, $\varnothing$, vessel.$n.02$, $\varnothing$$\rangle$&
$\langle$$<$PERSON$>$1, settle-36.1.2, $\varnothing$, indicator.$n.03$, indicator.$n.03$  $\rangle$ ; 
$\langle$music.$n.01$, escape-51.1-1, from, $\varnothing$, $\varnothing$$\rangle$;
$\langle$$<$PRP$>$, discover-84, to, run-51.3.2, progenitor.$n.01$$\rangle$&
The $<$ORG$>$0 can not scan the vessel.$n.02$ of the $<$VESSEL$>$0.
$<$PERSON$>$1 decides to be a little person.$n.01$ at the structure.$n.01$.
the music.$n.01$ arrives.
$<$PRP$>$ finds a lonely person.$n.01$ on the upper one of the craft.$n.02$ which is not a personal\_letter.$n.01$ but does not respond to hails .
&
The Jabba the Hutt can not scan the bareboat of the Uss Lakota.
O Yani decides to be a little mailer at the airdock.
The Music arrives.
She finds a lonely mailer on the upper one of the bareboat which is not a love letter but does not respond to hails.
\\\hline
Boba Fett has just chased down another bounty, a Rodian art dealer who sold fake works to Gebbu the Hutt.&
$\langle$$<$PERSON$>$0, chase-51.6, $\varnothing$, bounty.$n.04$, $\varnothing$$\rangle$&
$\langle$$<$PERSON$>$0, chase-51.6, to, magnitude.$n.01$, $\varnothing$$\rangle$;
$\langle$magnitude.$n.01$, comprehend-87.2, off, craft.$n.02$, magnitude.$n.01$$\rangle$;
$\langle$$<$PERSON$>$2, amuse-31.1, off, $\varnothing$, $\varnothing$$\rangle$;
$\langle$$<$PERSON$>$2, discover-84, off, change\_of\_integrity.$n.01$, $\varnothing$$\rangle$&
$<$PERSON$>$0 enters the bounty.$n.04$ and tells $<$PRP$>$.
$<$PERSON$>$0 attaches the explosive.$a.01$ to the person.$n.01$ who is trying to fix the device$.n.01$.
the magnitude.n.01 doesn't know the craft.$n.02$ off the craft.$n.02$.
$<$PERSON$>$2 is surprised when $<$PRP$>$ learns that the person.$n.01$ is actually $<$PERSON$>$7.
$<$PERSON$>$2 sees the change\_of\_integrity.$n.01$ and tells $<$PRP$>$.
&
Boba Fett enters the bounty and tells it.
Boba Fett attaches the explosive to the peer who is trying to fix the toy.
The multiplicity doesn't know the bounty off the bounty.
Dark Jedi Lomi Plo is surprised when it learns that the peer is actually Mrs Conners.
Dark Jedi Lomi Plo sees the combination off the Orbs and tells them.
\\
\end{tabular}
\end{table*}
\begin{table}[t]
\caption{Test set perplexity, BLEU, \& ROUGE (F1) scores, with average sentence lengths %, thresholds, and utilization percentages 
for event-to-sentence models.}

\label{table:results}
\footnotesize
\centering
\begin{tabular}{l|l|l|l|l}
{\bf Model}                & {\bf Perplexity} & {\bf BLEU} & {\bf ROUGE}& {\bf Length} \\ \hline
RetEdit              &  71.354 & 0.041 & {\bf 11.25} & 9.27 \\
Templates            & 203.629 &  0.0034 & 6.21 & 5.43 \\
Monte Carlo          & 71.385  & 0.0453 & 10.01 &7.91 \\
FSM & 104.775 & 0.0125 & 1.29 &10.98 \\
Seq2seq & 83.410 & 0.040 & 10.66&6.59  \\
RetEdit+MC & 72.441 & 0.0468 & 10.97 &9.41  \\
Templ.+MC & 79.295 & 0.0409 & 10.10 &6.92  \\
Templ.+FSM & 79.238 & 0.0296 & 6.36 &9.09 \\
RE+Tmpl.+MC & 73.637 & 0.0462 & 10.96 &9.35  \\
Full Ensemble             & {\bf 70.179}  & {\bf 0.0481} & 11.18 &9.22
\end{tabular}
\end{table}

\begin{table*}[]
\caption{Utilization percentages for each model combination on both events from the test set and from the full pipeline.}
\label{table:util}
\footnotesize
\centering
\begin{tabular}{l|l|l|l|l|l|l|l|l|l|l}
\multicolumn{1}{c|}{} & \multicolumn{2}{c|}{\textbf{RetEdit}}                     & \multicolumn{2}{c|}{\textbf{Templates}}                   & \multicolumn{2}{c|}{\textbf{Monte Carlo}}                 & \multicolumn{2}{c|}{\textbf{FSM}}                         & \multicolumn{2}{c}{\textbf{Seq2seq}}                     \\
                       & \multicolumn{1}{c|}{Test} & \multicolumn{1}{c|}{Pipeline} & \multicolumn{1}{c|}{Test} & \multicolumn{1}{c|}{Pipeline} & \multicolumn{1}{c|}{Test} & \multicolumn{1}{c|}{Pipeline} & \multicolumn{1}{c|}{Test} & \multicolumn{1}{c|}{Pipeline} & \multicolumn{1}{c|}{Test} & \multicolumn{1}{c}{Pipeline} \\ \hline
RetEdit+MC             & 82.58                     & 31.74                         &       -                    & -                             & 9.95                      & 48.4                          & -                         & -                             & 7.46                      & 19.86                         \\
Templates+MC           & -                         & -                             & 6.14                      & 5.48                          & 65.7                      & 66.67                         & -                         & -                             & 28.16                     & 27.85                         \\
Templates+FSM          & -                         & -                             & 6.14                      & 5.48                          & -                         & -                             & 56.77                     & 32.65                         & 37.09                     & 61.87                         \\
RetEdit+Templates+MC   & 82.58                     & 31.74                         & 1.49                      & 3.88                          & 9.1                       & 45.21                         & -                         & -                             & 6.82                      & 19.18                         \\
Full Ensemble          & 94.91                     & 55.71                         & 0.22                      & 0.91                          & 4.29                      & 41.10                         & 0.15                      & 0.68                          & 0.43                      & 1.60                         
\end{tabular}
\end{table*}
We perform two sets of experiments, one set evaluating our models on the event-to-sentence problem by itself, and another set intended to evaluate the full storytelling pipeline.
Each of the models in the event-to-sentence ensemble are trained on the training set in the sci-fi corpus.
The training details for each of the models are as described above.
All of the models in the ensemble slot-fill the verb automatically---filling a VerbNet class with a verb of appropriate conjugation---except for the sentence templating model which does verb slot-filling during post-processing. 
%---as part of the slot filling process.

After the models are trained, we pick the cascading thresholds for the ensemble by running the validation set through each of the models and generating confidence scores.
This is done by running a grid search through a limited set of thresholds such that the overall BLEU-4 score~\cite{papineni2002} of the generated sentences in the validation set is maximized.
These thresholds are then frozen when running the final set of evaluations on the test set.
For the baseline sequence-to-sequence method, we decode our output with a beam size of 5.
We report perplexity, BLEU-4, and ROUGE-4 scores, comparing against the gold standard from the test set.
\begin{equation}
  Perplexity = 2^{-\sum_{x}p(x)\log_2{p(x)}}
\end{equation}
\noindent
where $x$ is a token in the text, and
\begin{equation}
  p(x) = \frac{count(x)}{\sum_{v\in V}count(V)}
\end{equation}
where $V$ is the vocabulary.
%BLEU-4 is a metric commonly used in machine translation and calculates n-gram overlap---where $n$$=$$4$ here---between a series of gold standard ``references'' and predicted sentences. 
Our BLEU-4 scores are naturally low (where higher is better) because of the creative nature of the task---good sentences may not use any of the ground-truth $n$-grams. 
Even though we frame Event2Sentence as a translation task, BLEU-4 and ROUGE-4 are not reliable metrics for creative generation tasks.

The first experiment takes plots in the in the test set, eventifies them, and then uses our event-to-sentence ensemble to convert them back to sentences.
In addition to using the full ensemble, we further experiment with using different combinations of models along the spectrum between maintaining coherence and being interesting.
We then evaluate the generated sentences, using the original sentences from the test set as a gold standard.

The second experiment uses event sequences generated by an event-to-event system such as \citeauthor{Tambwekar2019}~\shortcite{Tambwekar2019} and is designed to demonstrate how our system integrates into the larger pipeline described in Figure~\ref{fig:pipeline}.
We then transform these generated event sequences into generalized sentences using both the ensemble and the baseline sequence-to-sequence approach.
As the last step, the generalized sentences are passed into the ``slot filler'' (see Figure~\ref{fig:pipeline}) such that the categories are filled.
As the story goes on, the ``memory'' maintains a dynamic graph that keeps track of what entities (e.g. people, items) are mentioned at which event and what their tag was (e.g. $<$PERSON$>$5, Synset(`instrument.n.01'))
When the slot filler sees a new sentence, it first tries to see if it can fill it in with an entity it as seen before.
This includes if the current Synset it is looking at is a descendent of a Synset already stored in memory.
If a new word has to be selected, named entities are randomly chosen from a list collected from the original science fiction corpus, with entities paired with their respective tags (PERSON, ORG, NUMBER, etc.).
Synsets are selected by finding a descendent 1 or 2 levels down.
The word is currently selected randomly, but this will soon be improved by the addition of a language model guiding it.
To fill a pronoun ($<$PRP$>$), the slot filler refers to the memory to select a recently-mentioned entity.
Person names are run through US Census data to determine the ``gender'' of the name in order to select an appropriate pronoun.
If no pronoun can be found, it defaults to {\em they}/{\em them}, and if no previous entity can be found, it defaults to {\em it}.
Organizations are always {\em they}.
For the purpose of this study, stories that came from the same events (story pairs across both conditions) were filled with the same entities.

Once the sentences from both experiments were complete, we conducted a human participant study on Amazon Mechanical Turk.
Each participant was presented a single story and given the list of 5-point Likert scale questions, validated by Purdy et al.~\shortcite{purdy2018} and used by Tambwekar et al.~\shortcite{Tambwekar2019} in their evaluation.
We exclude categories assessing the long-term coherence of a story as these categories are designed to evaluate an event-to-event system and not event-to-sentence, which is conditioned to map an event to a single sentence at a time.
Participants were also asked to provide a summary of the story and which of the attributes from the Likert questions thought to be most important for stories.
If the participants' English was not deemed fluent enough in the open-ended questions, their data was discarded.
This left us with 64 in the ensemble and 58 in the baseline condition.

%%%%%%%%%%%%%%%%%%%%%%%%%%%%%%%%%%%

\section{Results and Discussion}

\begin{figure}[t]
  \centering
  %\fbox{\rule[-.5cm]{0cm}{4cm} \rule[-.5cm]{4cm}{0cm}}
  \includegraphics[width=0.75\linewidth]{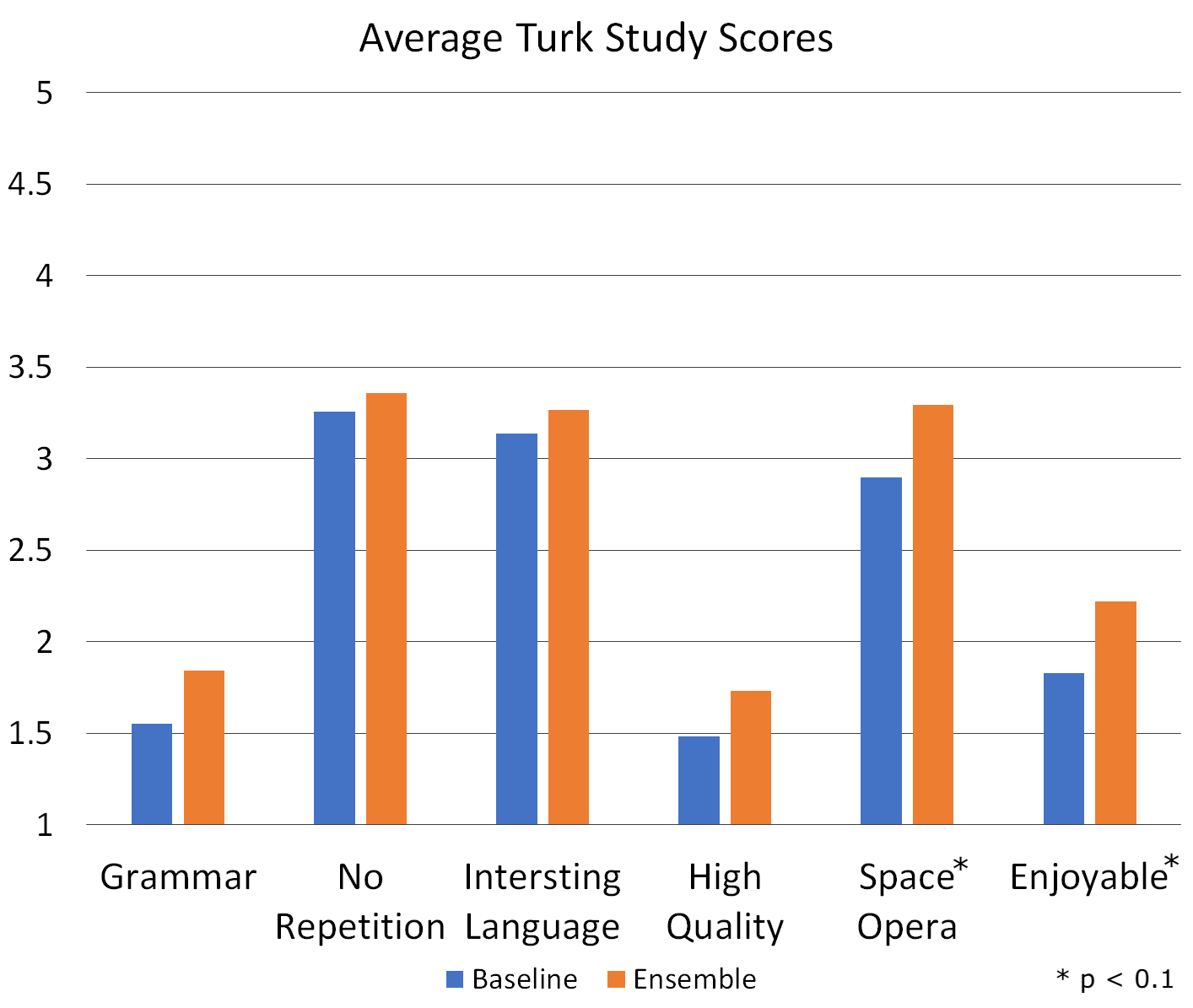}
  \caption{Human participant study results, where a higher score is better (scale of 1-5). Confidence values are 00.29 and 0.32, for genre and enjoyability respectively; $\alpha$=$0.1$. The confidence values for other metrics lie between 0.27-0.35.}
  \label{fig:humans}
\end{figure}

%Our results are presented in the form of two tables and a figure.
Table~\ref{table:results} shows the perplexity, BLEU-4 scores, ROUGE-4 scores, and average sentence length for event-to-sentence on the testing set for each of the models, ensemble, and baseline.
Note that some of the models, such as the sentence templates, make no use of gold standard sentences and are thus poorly optimized to use perplexity, BLEU, and ROUGE scores.
In addition to running each model in the ensemble individually, we experiment with multiple combinations of the models to assess which combination makes the most effective ensemble.
%All of the models used the same ordering as the full ensemble.
The full ensemble performs better than any of the individual models with regard to perplexity, as it is designed to combine the models such that each of their weaknesses is minimized.
The average sentence length metric highlights the differences between the models, with the templates producing the shortest sentences and the finite state machine taking longer to generate sentences due to the constraints it needs to satisfy.
%Table~\ref{table:results2} shows the confidence thresholds after tuning the ensemble.
%The RetEdit and sentence template models need 80\% confidence in their results, or the next model in the cascade is tried.

We also noted how often each model in the ensemble is used, shown in Table~\ref{table:util}, when generating sentences from the eventified testing corpus or from the event-to-event model within the pipeline, across different combinations of ensembles.
%and the utilization percentages for the models on the event-to-sentence testing set and the events generated by the event-to-event system.
Utilization percentages show us how often each model was picked in the respective ensembles based on the corresponding confidence score thresholds.
RedEdit was heavily used on the test set, likely due the train and test sets having a similar distribution of data.
On the pipeline events, RetEdit is used much less---events generated by event-to-event are often very different from those in the training set.
A majority of the events that fall through RetEdit are caught by our Monte Carlo beam search, irrespective of the fact that RetEdit---and sentence templates---are most likely to honor the event tokens.
This is partially due to the fact that satisfying the constraint of maintaining the events tokens makes it difficult for these models to meet the required threshold.
The small portion of remaining events are transformed using the templates and the finite state machine.

Table~\ref{table:examples-e2s} shows examples of generalized sentence outputs of each of the event-to-sentence models, illustrating some of the trends we alluded to in Section 3.
%Some trends are evident.
%We observe a few trends as we have previously predicted.
Retrieve-and-edit focuses on semantics at the expense of sentence quality.
The sentence templates produce output that matches the input event but is very formulaic.
Monte Carlo generates entertaining and grammatically-correct sentences but occasionally loses the semantics of the input event.
The finite state machine attempts to achieve a balance between semantics and generating entertaining output, however it sometimes fails to produce an output given the constraints of the state machine itself.
All of these can be compared to the original next event from the testing set.
We also provide examples of the entire pipeline in Table~\ref{table:examples-pipe}, which 
demonstrates our ensemble's ability to work with an existing plot generator.

For the human participants study comparing a seq-to-seq baseline to our full ensemble (Figure~\ref{fig:humans}), most metrics were similar in score, which is understandable given that both conditions produced stories that were at times confusing.
However, the ensemble consistently outperformed the baseline in terms of quality, maintaining the genre (space opera), grammar, and enjoyablity.
Enjoyability and genre were significant at $p<.10$ using a two-tailed independent t-test.

\section{Conclusions}

Event representations improve the performance of plot generation and allow for planning toward plot points. 
However, they are unreadable and abstract, needing to be translated into syntactically- and semantically-sound sentences that can both keep the meaning of the original event and be an interesting continuation of the story.
%We present an ensemble of four event-to-sentence models that act as part of an overall plot generation pipeline.
We present an ensemble of four event-to-sentence models, in addition to a simple beam search model, that balance between these two problems.
Each of the models in the ensemble is calibrated toward different points in the spectrum between the two issues and are thus designed to cover each other's weaknesses.
The ensemble is integrated into a full story generation pipeline, 
%based off of 
%the partial pipeline of 
%Martin et al. (\citeyear{Martin2018}), 
%taking an input sentence and returning the next sentence of the story, 
demonstrating that our ensemble can work with generated events drawn from a realistic distribution.

\fontsize{9.0pt}{10.0pt}
\selectfont
\bibliography{aaai20.bib}
\bibliographystyle{aaai}

\end{document}